\documentclass[conference]{IEEEtran}

\ifCLASSINFOpdf
\else
\fi

\usepackage{graphicx}
\graphicspath {{./figure/}}
\usepackage{hyperref}

\begin{document}


\title{A*SLAM: A Dual Fisheye Stereo Edge SLAM}
\author{\IEEEauthorblockN{Guoxuan Zhang} 
\IEEEauthorblockA{Astar.ai, Inc.\\
Email: \href{mailto:support@astar.ai}{support@astar.ai}, Web: \href{https://astar.ai}{https://astar.ai}}}

\maketitle


\begin{abstract}
This paper proposes an A*SLAM system that features combining two sets of fisheye stereo cameras and taking the image edge as the SLAM features. The dual fisheye stereo camera sets cover the full environmental view of the SLAM system. From each fisheye stereo image pair, a panorama depth image can be directly extracted for initializing the SLAM feature. The edge feature is an illumination invariant feature. The paper presents a method of the edge-based simultaneous localization and mapping process using both the normal and inverted images interchangeably.
\end{abstract}


\IEEEpeerreviewmaketitle


\section{Introduction}

In a complicated environment, all self-driving mobile systems need to equip with the vision sensors. Therefore, if we can implement a SLAM and navigation system purely depends on the vision sensors, then we can save a big amount of cost by excluding the expensive laser and/or LiDAR sensors.

A SLAM system is not just for the mapping and localization purpose, it also needs to support the robust navigation functionality. In a practical application, the issue of \textit{can} SLAM should be replaced with \textit{robust} SLAM nowadays. This short paper poses the problem of what are the necessary components for a robust visual SLAM system, and then introduce the A*SLAM system which meets all of these conditions.

\begin{figure} [!b]
    \centering
    \footnotesize
    \hspace{0.3cm} \includegraphics[width=4.5cm ]{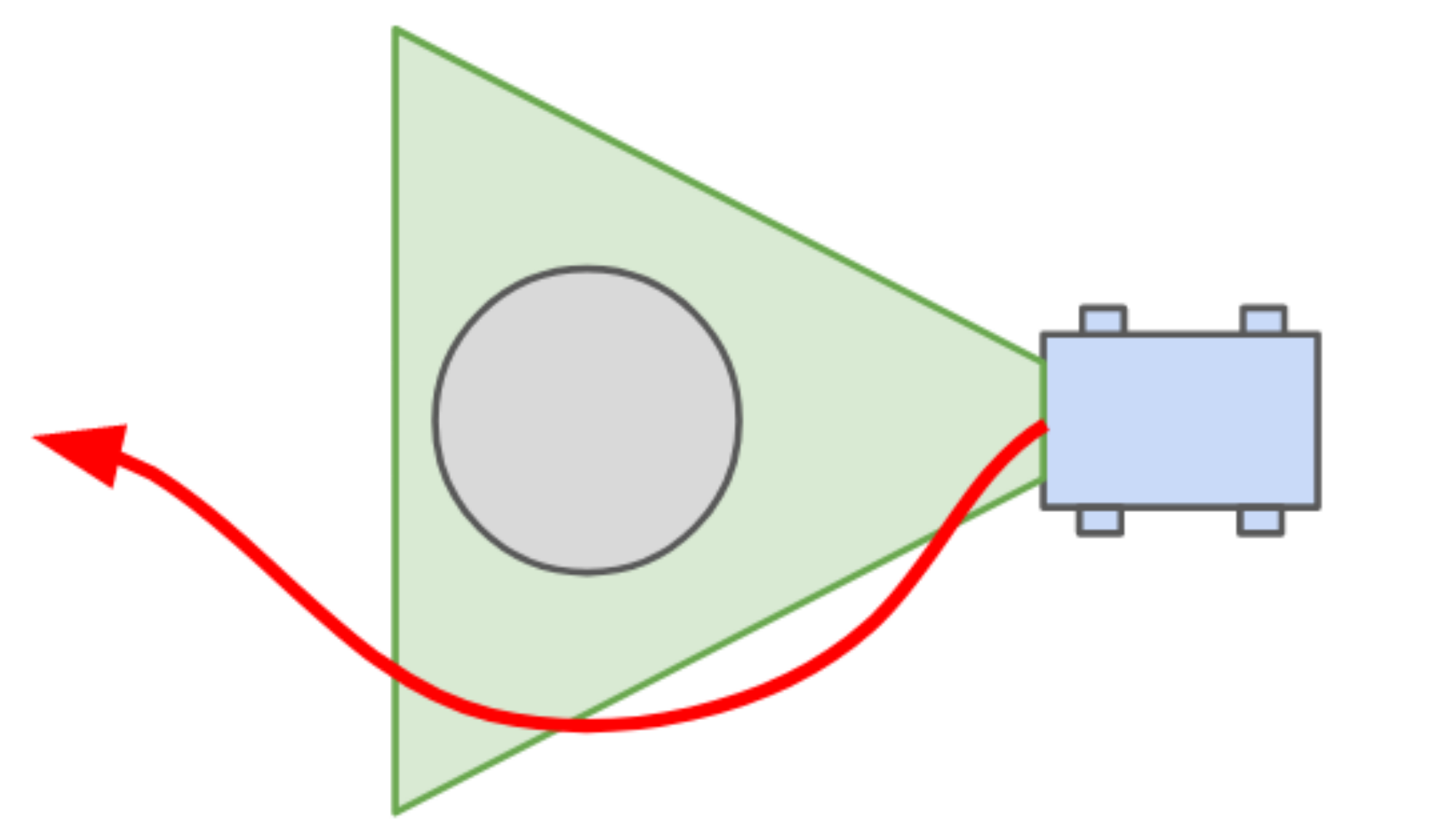} \hspace{0.2cm}
    \includegraphics[width=2.9cm ]{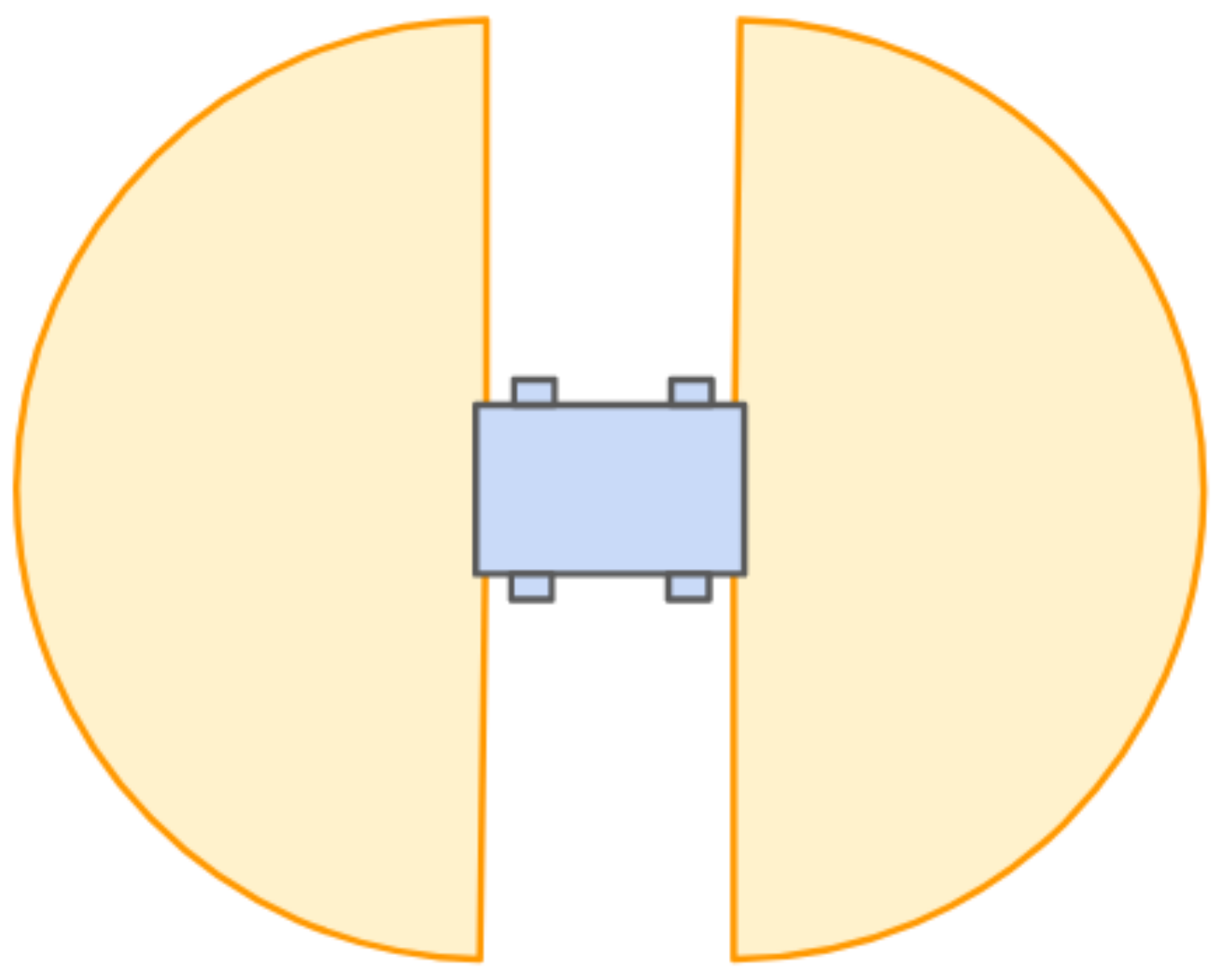} \\
    (a) Path planning for obstacle avoidance \hspace{0.4cm} (b) Full view
    \caption{(a) In a navigation scenario, the robot is failed to verify the planned path because of a limited camera FoV. (b) A full surrounding view provided by dual fisheye camera configuration.}
    \label{fg::obstacle}
\end{figure}

\section{Requisites for Robust Visual SLAM}

A robust visual SLAM system needs to satisfy at least the following four conditions.

\subsection{Stereo Vision}

A monocular SLAM can only build a map up to scale. A scale map is insufficient for a mobile system to perform meaningful work in the physical world. The motion control is based on the metric information, and the movements are represented by the metric unit. A stereo vision can add metric measurements to the map based on the baseline distance of the binocular imaging.

Another benefit of the stereo vision is the detection of depth value for each pixel by fast image matching along the predefined epipolar line. The depth information can be used for the initialization of the SLAM feature, and can also be used for obstacle detection in a navigation system.

\begin{figure} [t]
    \centering
    \includegraphics[width=6cm]{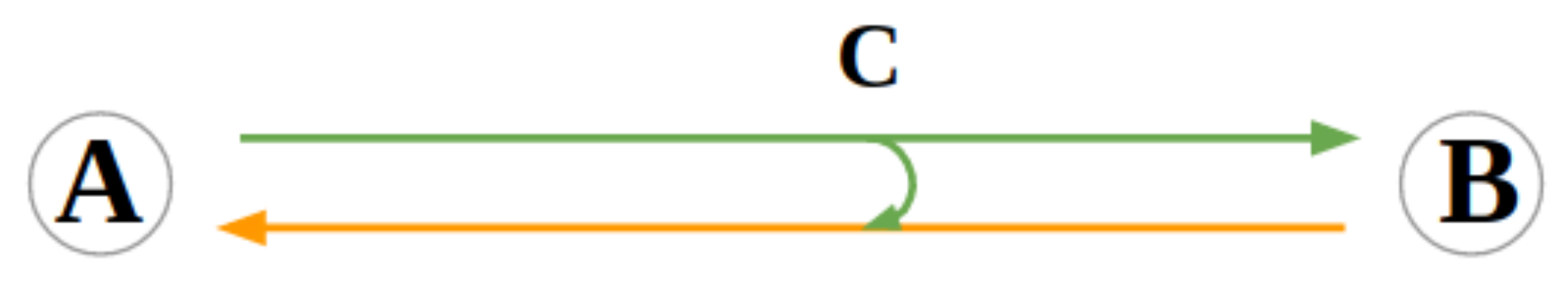}
    \caption{To map an area between A and B, the partial view camera needs to travel twice to form a complete map. When the robot makes a turn at the midpoint C, the map switching tends to be failed due to the mapping error.}
    \label{fg::two_path}
\end{figure}

\subsection{Wide Field of View}

Wide field of view (FoV) means possible of integrating more spatial information in a SLAM computation. Seeing wider and more can usually improve the accuracy and the robustness of the feature tracking. Furthermore, the SLAM features near outer FoV can be easily converged because these features can utilize big parallax existing at the outer FoV area.

In a navigation scenario, a limited FoV will degrade the path planning performance. Figure \ref{fg::obstacle}(a) shows a robot (blue) is blocked by an obstacle (gray). After the robot planned a path (red), however, because the camera FoV (green) is not wide enough to verify the planned path, so the robot cannot convert this path to motion immediately. 

Adopting the fisheye camera is considered as the simplest way to provide a wide FoV to a visual SLAM system. 

\subsection{Full View}

The concept of full view is an extension of wide FoV but it requires using more than one camera. Figure \ref{fg::obstacle}(b) shows a full surrounding view configuration using two fisheye cameras. Compared to a partial camera view, the full view provides even more robustness to the SLAM system. In a full view system, the occlusion problem becomes much more trivial than the single camera system.

Beyond the above-mentioned advantages, the main reason for using the full view configuration is the full view can generate a complete map. A complete map means a single scan can gather all 360-degree information into a map, hence, the same area does not need any further scan. As shown in Fig. \ref{fg::two_path}, to map an area between A and B, a partial view camera needs to scan the area twice. After mapping, there exist two independent maps A-B and B-A. When a robot started from A and then made a turn at midpoint C, there is no guarantee the robot can smoothly switch from map A-B to map B-A due to the inevitable mapping error. For practical use, all visual SLAM systems should be comprised of the full view configuration in the future.

\begin{figure} [t]
    \centering
    \includegraphics[width=7cm]{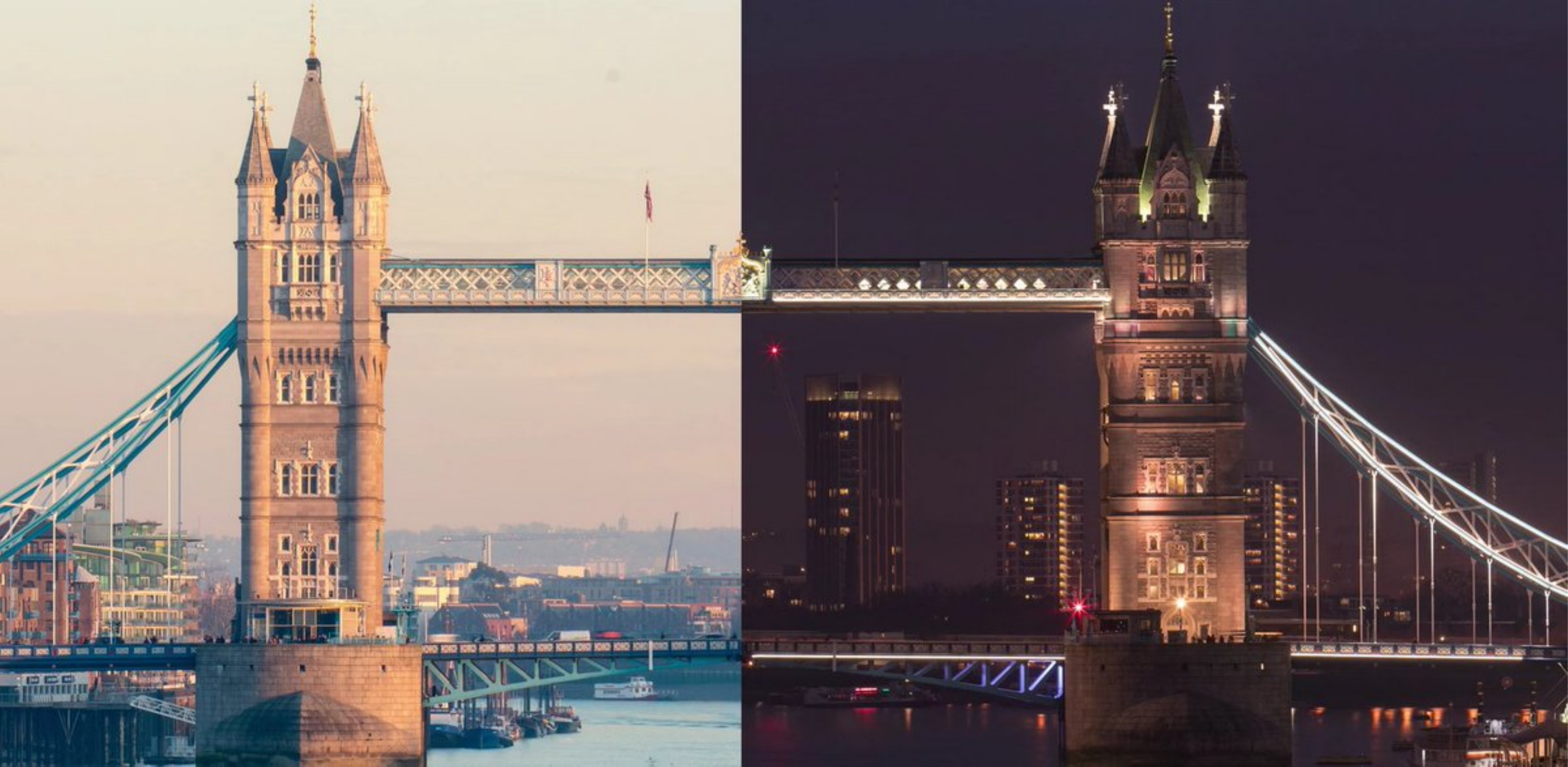}
    \caption{Edges are illumination invariant features. In this image, the edges from the object contours preserve the shape information even under severe light changing conditions (\textcopyright \href{https://www.matel.tv/}{matel.tv}).}
    \label{fg::edge}
\end{figure}

\begin{figure} [b]
    \centering
    \includegraphics[width=7cm]{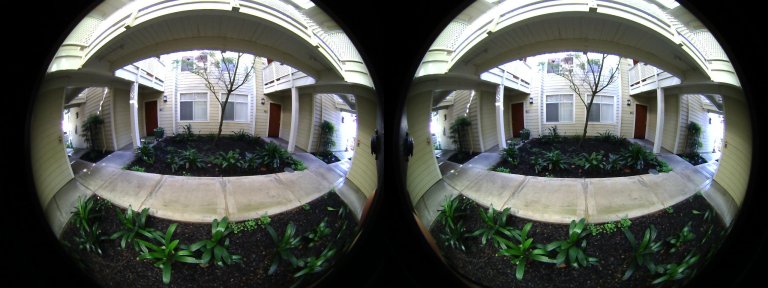} \\ \vspace{0.05cm}
    \includegraphics[width=7cm]{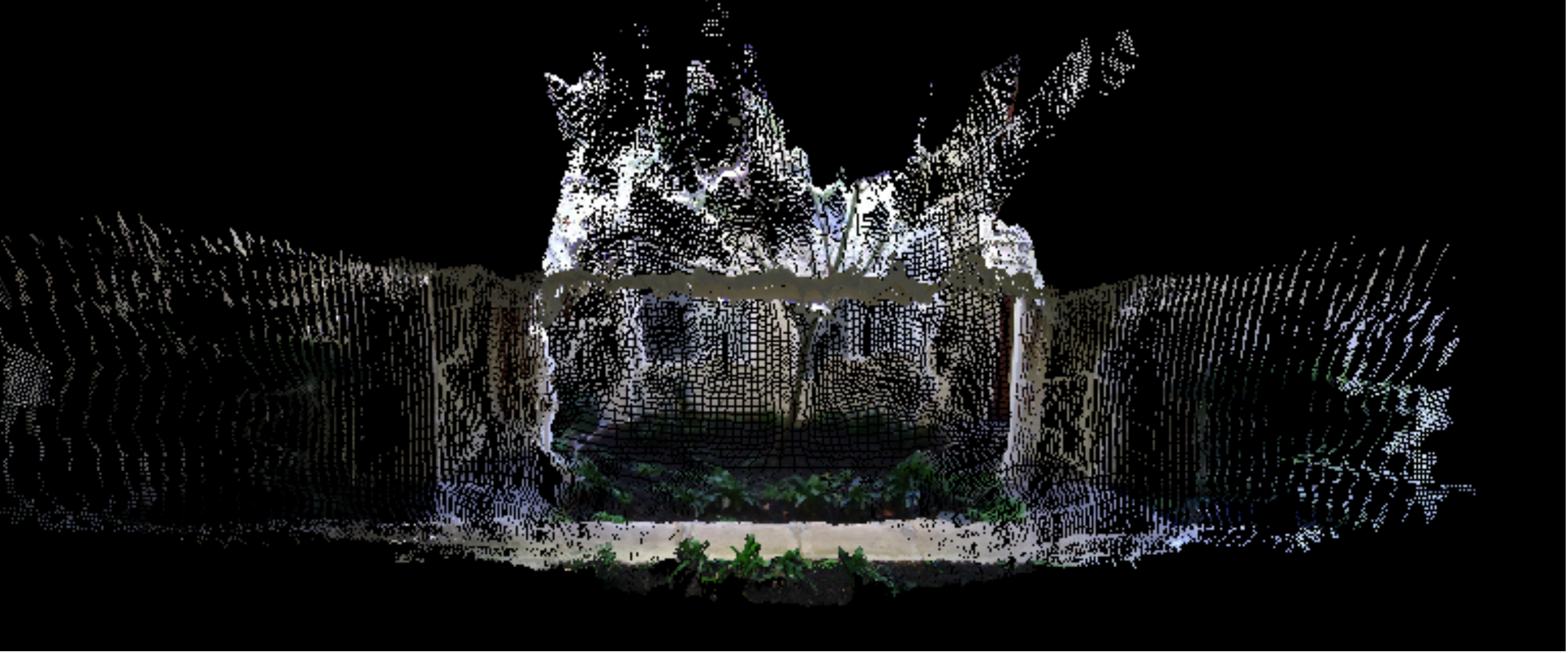}
    \caption{A panorama depth image can be directly extracted from a pair of fisheye stereo images. The depth information is used for initializing the SLAM features and the obstacle detection in a navigation system.}
    \label{fg::pdi}
\end{figure}

\subsection{Illunimation Invariant Feature}

A SLAM map is a static one, it can only include a specific lighting condition when the map is built. By comparison, the environment is a dynamic one, the visual appearance suffers from the frequent change of lightings, weathers, and seasons. Existing popular SLAM methods, like \cite{orb} and \cite{dso}, are all using the pixel value to compose the SLAM features. Accordingly, the performance of tracking the SLAM feature is greatly downgraded when the lighting condition is different from the registration time. 

Edge is an illumination invariant feature. Edges are formed by the abrupt color transitions on a surface or the intersection of heterogeneous geometric structures in 3D space. As shown in Fig. \ref{fg::edge}, the edges can be robustly detected at the same location even under the severe change of the lighting conditions. Dealing edges as a pure geometric substance, without the supporting of the pixel values, is a very challenging problem in the SLAM community. Now we are proud to announce the problem of using the \textit{edge} as the SLAM feature is solved.


\section{A*SLAM}

A*SLAM system satisfies all previously described four conditions of a robust visual SLAM system. A*SLAM system features combining two sets of fisheye stereo cameras and taking the image edge as the SLAM features. 

Among the dual stereo camera sets, one is looking forward and another one is looking backward. Each stereo camera is equipped with a pair of 180-degree fisheye lenses. In this way, the A*SLAM system is able to cover a full environmental view. The fisheye image can also be used to generate a wide-angle depth image. Figure \ref{fg::pdi} shows an example of using our developed {CaliCam}\textsuperscript{\textregistered} \cite{calicam} stereo camera to generate a panorama depth image \cite{pdi}. 

\begin{figure} [t]
    \centering
    \includegraphics[width=8cm]{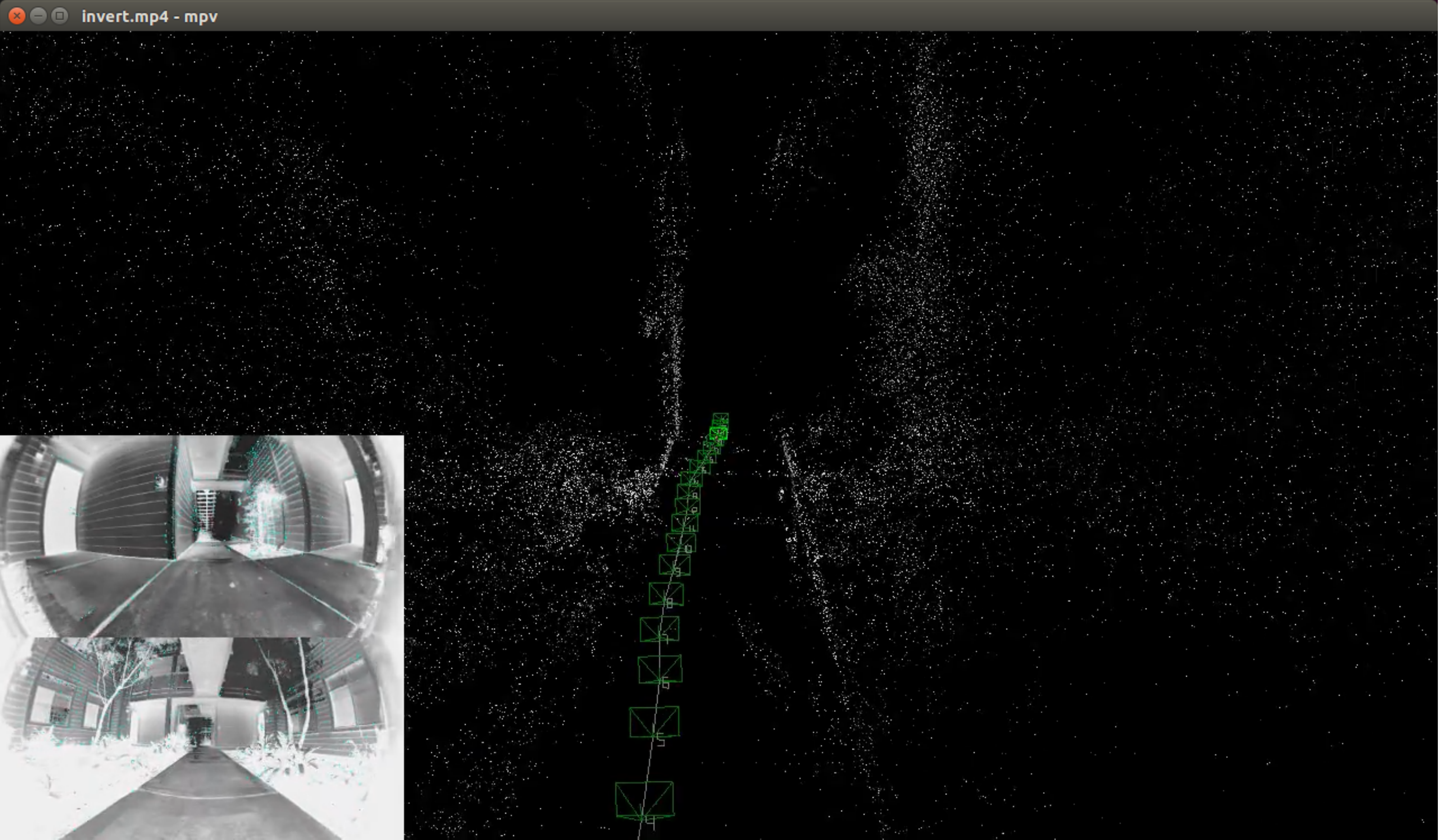}
    \caption{The bottom left part shows two inverted images taken from the front camera and the back camera, respectively. The localization and mapping process can be conducted with both the normal and inverted images interchangeably in the A*SLAM system.}
    \label{fg::aslam}
\end{figure}

By abstracting the image edge as a pure geometric substance, A*SLAM system enables a simultaneous localization and mapping process based on both the normal and inverted images interchangeably, as shown in Fig. \ref{fg::aslam} and a demonstration video clip \cite{youtube}. 


\section{Conclusion}

To the best of our knowledge, A*SLAM is the only system in the world to embrace all the challenging requisites of a robust visual SLAM system. We are now actively looking for all potential domains to apply our robust A*SLAM system for maximizing the SLAM and navigational performance.

\end{document}